%% file: main.tex
\documentclass[10pt,twocolumn,letterpaper]{article}
\usepackage[T1]{fontenc}
\usepackage{authblk}
\usepackage[T1]{fontenc}
\usepackage[utf8]{inputenc}

\usepackage[pagenumbers]{cvpr} 

\usepackage{graphicx}
\def\shrug{\texttt{\raisebox{0.75em}{\char`\_}\char`\\\char`\_\kern-0.5ex(\kern-0.25ex\raisebox{0.25ex}{\rotatebox{45}{\raisebox{-.75ex}"\kern-1.5ex\rotatebox{-90})}}\kern-0.5ex)\kern-0.5ex\char`\_/\raisebox{0.75em}{\char`\_}}}

\input{preamble}

\definecolor{cvprblue}{rgb}{0.21,0.49,0.74}
\usepackage[pagebackref,breaklinks,colorlinks,allcolors=cvprblue]{hyperref}

\title{Spatial Representation Learning Beyond Pixels: Unifying Raster Data and Vector Semantics for Human-Centric Geospatial Foundation Models}

\makeatletter \renewcommand\AB@authnote[1]{} \makeatother

\author{Steffen Knoblauch$^{1}$ \; Hao Li$^{2}$  \; Gengchen Mai$^{3}$ \quad Konstantin Klemmer$^{4}$ \; \\ Song Gao $^{5}$\; WenWen Li$^{6}$}

\affil[ ]{$^{1}$Heidelberg University, $^{2}$National University of Singapore, $^{3}$University of Texas at Austin, $^{4}$University College London, \newline $^{5}$University of Wisconsin-Madison, $^{6}$Arizona State University}

\affil[ ]{\texttt{steffen.knoblauch@uni-heidelberg.de, hao.li@nus.edu.sg, gengchen.mai@austin.utexas.edu}}
\affil[ ]{\texttt{song.gao@wisc.edu, wenwen@asu.edu}}

\begin{document}
\twocolumn[{%
\renewcommand\twocolumn[1][]{#1}%
\maketitle
\centering
\includegraphics[width=1\linewidth]{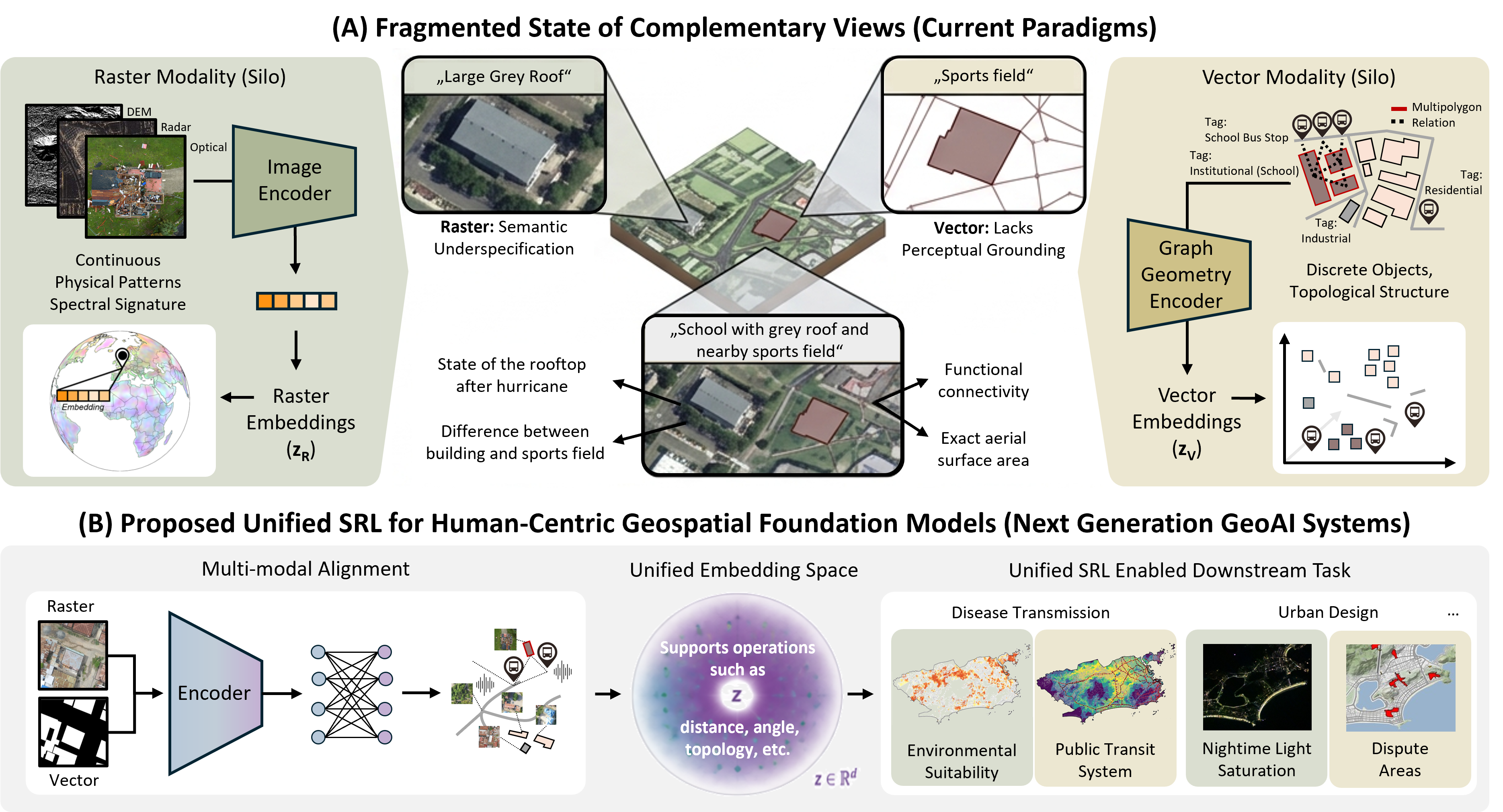} 
\captionof{figure}{From Silos to Synthesis: Toward human-centric geospatial foundation models through unified SRL across vector and raster data. \label{fig:Unified_embedding_space}
}
\vspace{2.5em}
}]

\begin{abstract}
Earth Observation (EO) has fundamentally transformed the monitoring of environmental processes and human activities up to planetary scale. Recent advances in self-supervised learning have given rise to Earth Observation Foundation Models (EOFMs), which leverage petabyte-scale archives of unlabeled EO data to learn transferable representations across a wide range of downstream geospatial tasks, including land cover mapping, disaster response, and climate monitoring. Despite these advances, current EOFMs remain largely confined to raster modalities, overlooking the rich, structured information encoded in openly-accessible vector data sources such as OpenStreetMap and Overture. Vector data provides explicit and compact representations of geographic entities, including geometry, topology, and semantic relationships, offering critical contextual signals that are often ambiguous or inaccessible in imagery alone. Raster and vector data thus represent complementary views of geographic space: raster data captures continuous physical and spectral patterns, while vector data encodes discrete objects and their relational structure and often represents more of the human rather than the physical systems (e.g. social or demographic data). However, existing geospatial representation learning paradigms treat these modalities in isolation, relying on imperfect and often lossy transformations to bridge them. This perspective paper calls for a paradigm shift toward joint Spatial Representation Learning (SRL) in an unified embedding space that integrate raster perception with vector-based reasoning. Building on emerging efforts in multimodal geospatial learning, we highlight conceptual foundations, technical challenges, and promising directions for aligning heterogeneous spatial data sources. We contend that such integration is essential for developing next-generation geospatial AI systems capable of more accurate, interpretable, and semantically grounded understanding of the Earth.
\end{abstract}

\section{Introduction}
Earth observation (EO) through satellite imagery has transformed our ability to monitor global environmental processes and human activities. The recent emergence of self-supervised foundation models marks a pivotal advance in geospatial artificial intelligence (GeoAI) and earth observations~\citep{mai2025towards}, enabling scalable representation learning from vast unlabeled archives of overhead imagery. These models incorporate specialized architectures that account for the unique properties of geospatial data, including multi-resolution observations, temporal sequences, and multi-spectral channels \cite{cong2022satmae,Reed.2022}. Through pretraining on petabyte-scale datasets, they produce robust feature representations that transfer effectively to diverse downstream applications such as land cover classification, disaster assessment, crop field delineation, and climate monitoring \cite{Bodnar.2025, Dollinger.2025}.

Despite their success, current Earth Observation Foundation Models (EOFMs), typically trained on Earth observation imagery such as satellite or aerial raster data, and broader Geospatial Foundation Models (GeoFMs), which may additionally encompass other geospatial modalities including vector and structured data, remain largely confined to raster-based representations~\citep{Klemmer.2025}, overlooking the rich, structured geospatial knowledge encoded in openly-accessible vector data. Vector data such as OpenStreetMap and  Overture capture explicit spatial entities, including buildings, roads, and land use parcels, along with their geometry, topology, and semantic relationships \cite{Bai.2025,Bai.2025b}. These representations reflect human interpretation and organization of space, providing contextual information that is often ambiguous or inaccessible from imagery alone, such as functional distinctions between visually similar structures or connectivity within infrastructure networks \cite{VargasMunoz.2021}.

Raster and vector data thus embody fundamentally complementary abstractions of geographic reality. Raster representations capture continuous physical and spectral patterns but approximate object boundaries and obscure relational structure \cite{Reed.2022}. Vector representations preserve precise geometry, topology, and semantic identity but lack dense physical detail and environmental context \cite{Mai.2024b}. Yet, current GeoAI development often treats these modalities as separate silos, linked only through lossy transformations such as rasterization or vector extraction, and feature engineering, which hinder structural fidelity and limit cross-modality reasoning \cite{Bai.2025b, Gomes.2025} (cf. Fig. \ref{fig:Unified_embedding_space}).

This fragmentation constrains the development of geospatial models capable of holistic understanding. Many real-world applications, from urban analysis to climate impact assessment—demand simultaneous reasoning over physical observations and human, defined spatial structure \cite{mai2025towards}. However, existing deep learning-based approaches cannot fully exploit these complementary signals, since these two paradigms require distinct ways for neural encoding and representations which are hard to align and integrate in a fine-grained manner\cite{Bai.2025}, e.g., alignment of one vector-based building polygon to its corresponding set of pixels in a satellite image. Notably, using raster or vector data alone has intrinsic limitations. While raster-based models remain semantically under-specified, vector-based methods lack contextual grounding \cite{Liu.2025b}. Combining both data models could provide a more comprehensive view of the physical world, as shown in Figure \ref{fig:Unified_embedding_space}, and allow for a deeper understanding of the interconnections between the physical and human processes shaping our planet.

We envision that addressing this limitation requires a shift toward joint SRL, in which raster and vector data (and their inter-dependencies) are learned within shared embedding spaces. Such representations would unify perceptual and structural information, enabling models to align visual patterns with semantic entities and their relationships. This perspective paper outlines the conceptual foundations, technical challenges, and emerging directions for such unified approaches, and positions joint raster–vector learning as a critical step toward next-generation, human-centric geospatial foundation models \cite{Mai.2024,Liu.2025b}.

\section{Conceptual Foundations of Spatial Representation Learning}
\subsection{Raster Representations and Imagery-based Approaches}
Over the last years, EO and Remote Sensing (RS) data have become abundant and available at the petabyte scale, providing almost real-time insights into natural and human dynamics on Earth. At the same time, computer vision and representation learning have seen major advances, especially w.r.t model pre-training and self-supervised learning, where general-purpose data representations were successfully obtained from ImageNet and, later on, domain-specific RS datasets \cite{hong2024spectralgpt, Reed.2022}. Remote sensing imagery captures geographic space as regular grids of discrete cells, constructing an ideal representation of continuous geographical layers such as elevation profiles, vegetation indices, and land cover information. By harnessing the power of massive EO data, advanced imagery-based EOFMs often employ self-supervised pretraining objectives tailored to geospatial characteristics \citep{Szwarcman.2025}. Advanced AI architectures, like Masked autoencoder (MAE), treat the model training as self-supervised reconstruction which randomly masks patches while maintaining multi-scale contextual coherence, results in latent feature sensitive to both local textures and global scene structure \cite{Feng.2025, Tseng.2025}. Moreover,  contrastive frameworks further align embeddings of spatially adjacent patches while repelling distant regions, embedding the spatial autocorrelation characteristic of RS imagery toward general-purposed SRL of EO data.

Recently, a growing body of work demonstrates the effectiveness of these strategies at scale. Models such as SkySense \cite{Zhang.2025}, SpectralGPT \cite{hong2024spectralgpt}, SatCLIP \cite{Klemmer.2025}, TerraFM \cite{Danish.2025}, CopernicusFM \cite{Wang.2025}, and Prithvi-EO-2.0 \cite{Szwarcman.2025} leverage multi-modal and multi-sensor EO data to generate powerful neural representations that perform robustly across geographic regions, sensor modalities, and downstream geospatial tasks. This landscape has been further expanded by the emergence of AlphaEarth \cite{Brown.2025}, which demonstrates large-scale integration of multi-sensor and multi-temporal observations into a unified Earth representation. Moreover, recent developments, for example THOR \cite{forgaard2026thor}, DOFA-CLIP \cite{Xiong.2025} and TerraMind \cite{Jakubik.2025}, further introduce compute-adaptive architectures and incorporate textual alignment to handle sensor heterogeneity and variable resolutions as well as to enable zero-shot generalization \cite{forgaard2026thor}. Beyond optical RS imagery, Earth system FMs (e.g., Prithvi WxC \cite{Schmude.2024} and Aurora \cite{Bodnar.2025}) extend the SRL paradigms to atmospheric and climate variables, demonstrating the potential and scalability of EOFMs across petabyte-scale datasets \cite{Wang.2024}.

Raster representations excel at scalable perception of environmental properties but face limitations stemming from their inherent grid structure. However, in the real-world, geospatial objects often have boundaries spanning across pixel-level approximation, and feature complex topological properties such as semantic association that can go far beyond from local pixel-based information can capture \citep{Mai.2024b}. All these characteristics render imagery-based SRL to be extremely effective for pattern recognition but fundamentally limited for complex geospatial tasks demanding precise geometric reasoning \cite{mai2023towards,Siampou.2024} or explicit object-level semantic understanding \citep{adimoolam2025pix2poly}.

\subsection{Vector Representations and Polymorphic-based Approaches}
To extract meaningful features from vector data for downstream tasks, many researchers either perform feature engineering based on domain knowledge, or convert spatial data from their original formats (e.g., points, polylines, and polygons) into formats that are easier for neural networks to handle (e.g., point clouds to voxels, or map vector files into raster image tiles \cite{balsebre2024city}.

As polymorphic vector data lacks the regular grid structure of pixels, it is considered as a more challenging data source (e.g., points, polylines, polygons, triangulated irregular networks, 3D LiDAR point clouds, 3D building models, etc.) for SRL to learn from. Fortunately, recent developments of SRL techniques start to develop new SRL methods for polymorphic vector data (e.g., point, line and polygon). Specifically, polymorphic-based SRL approaches map non-Euclidean vector geometries into low-dimensional, dense embedding vectors while preserving their spatial, structural, and topological properties \citep{Siampou.2024}.

For geotagged point data, advanced approaching such as location encoding based on sinusoid functions and spherical harmonics provide continuous representations of geographic position \cite{Ruwurm.2023}, which may be aggregated into arbitrary polygon geometries, although systematic studies on aggregating either location embeddings or global pixel-level embeddings such as Alpha Earth Foundations onto a polygon geometry remain scarce. Besides point data encoding, advances in global-scale spatial indexing, such as the S2 Geometry-based S2Vec \cite{Choudhury.2026}, further facilitate robust global-scale geographic feature encoding. Beyond point-level inputs, frameworks like GeoBind \cite{Dhakal.2024} and the Major TOM project \cite{Czerkawski.2024} provide high-dimensional abstractions for mapping individual geographic layers into shared latent spaces for the purposed of SRL. Recent geometry-aware approaches such as Poly2Vec \cite{Siampou.2024} and Geo2Vec \cite{Chu.2026} is able to encode points, shapes, and distances simultaneously while preserving geometric structure. More recently, multimodal FMs like AETHER \cite{Liu.2025b} start to emphasize the role of POIs as semantic anchors, leveraging multimodal alignment to link vector-based human activity patterns with imagery-based urban environment observation via RS and SVI. Recent evaluations also highlight the potential of Large Language Models (LLMs) in understanding geometries and topological spatial relations \cite{Ji.2025}. In addition, we observe that more object-based and region-level methods integrating visibility graphs, POI hierarchies, and urban functional structure through hierarchical graph learning \cite{Yu.2024}, also show great potential in extending the SRL scope into the untouched polymorphic vector dataset.

In short, there is a pressing need of advanced vector representations, via the SRL of polymorphic dataset, which can guarantee geometric fidelity (e.g., shape and distance), explicit topological structure (e.g., connectivity and morphology), and semantic interpretability (e.g., POI spatial association). This need comes not only from scientific perspective, but also from the practical consideration of geospatial reasoning and multimodal data integration. Herein, we envision a similar importance and benefit for advanced SRL methods from both raster and vector geospatial data, even within a unified framework.

\section{Toward Integrating Vector Semantics into Raster-Based Spatial Representation Learning}
Raster grids excel at representing continuous fields and visual patterns, but can approximate object boundaries and obscure topological relationships, depending on the spatial resolutions and scales. Vector primitives preserve exact geometry, connectivity, and semantic identity but lack the dense physical context available from satellite observations. Current GeoAI models usually treat these modalities as separate silos with distinct neural architectures for SRL, connected only through lossy, task-specific conversions that sacrifice structural fidelity \cite{Gomes.2025}. In the following, we will systematically examine the limitations of existing integration strategies, highlight emerging directions for native multimodal architectures, and argue for a unified representation learning paradigm that treats raster and vector data as coequal components within shared embedding spaces.

\subsection{The Imperative for Unified Spatial Representations}
Real-world geographic systems reflect the confluence of--often continuous--environmental spatial processes and--often discrete--human spatial organization. Atmospheric dynamics interact with transportation corridors, vegetation succession constrains parcel utilization, and functional urban districts emerge from the articulation of built form with topographic and climatic context. Comprehensive spatial representation, therefore, demands multi-scale processing across raster grids and vector object hierarchies, cross-modal alignment between pixel spaces and discrete entities, and collaborative pretraining regimes that harness complementary inductive biases from both data domains--and learn interactions between them.

\subsection{Current Integration Strategies: Limitations of Indirect Approaches}
A major challenge in raster and vector joint learning lies in the fundamental differences between raster and vector data representations. Raster data, such as satellite imagery, represent continuous spatial fields and naturally align with the inductive biases of deep learning architectures, particularly convolutional and transformer-based models designed to process dense grid structures. In contrast, vector data are inherently discrete, sparse, and structured, consisting of geometries, topological relationships, and semantic attributes. Because of these differences, designing effective SRL models for vector data remains challenging. As a result, vector-to-raster conversion is a common strategy that converts vector data into raster images before feeding it into neural networks. Such practice is usually adopted due to the lack of specialized approaches directly encoding vector data into the neural networks \cite{Mai.2024b}. As a traditional GIS approach, it is still widely used in many recent GeoAI studies, such as building/building group classification and generalization (converting vector building polygons into map images) \cite{zhou2024spagan}, air pollution forecasting (i.e., converting point-based air pollution measures into interpolated air pollution field) \cite{zhang2026eulerian}, and transportation model identification (converting GPS trajectories into trajectory images) \citep{ribeiro2025deep}. The drawbacks of such an approach are well-documented: 1) conversions can lead to substantial information loss of shape details and topological information, and losses based on vector-to-raster functions are, in most cases, irreversible since the conversion process is non-differentiable; 2) arbitrary scale decisions, e.g., the pixel size, can lead to a modifiable areal unit problem (MAUP), when the to-be-modeled process is represented differently depending on spatial scales or hierarchies.

\subsection{Toward Unified Spatial Representation Learning}
Emerging architectures demonstrate the viability of native multimodal processing that treats raster patches and vector objects as parallel token streams within shared transformer frameworks. Foundational vision papers articulate the necessity of joint modality processing for comprehensive geospatial artificial intelligence \cite{Mai.2024}. Projects such as SkyScript \cite{Wang.2023c} exemplify this shift, constructing large-scale vision-language datasets by using geo-coordinates to pair satellite imagery with OpenStreetMap semantics, thereby enabling the development of task-agnostic, zero-shot remote sensing models. However, the vector geometries are not explicitly considered in this dataset. 
Rose \cite{Bai.2025b} and GeoLink \cite{Bai.2025} overcome this problem by performing 
cross-modal self-supervised pretraining that aligns remote sensing representations with OpenStreetMap geometry neural representations via masked autoencoder and contrastive leanring objectives while preserving structural detail \cite{Bai.2025}. 
However, both Rose \cite{Bai.2025b} and GeoLink \cite{Bai.2025} ignore the geometry details of geographic objects (e.g., polylines and polygons) and abstract them into a single location, which leads to significant information loss. 
To facilitate raster and vector integration across disparate scales, recent research utilizes discrete global grid systems like S2 Geometry as a unifying spatial index. For instance, S2Vec \cite{Choudhury.2026} demonstrates how geographic feature vectors can be tokenized within such a hierarchical framework, allowing pixels and polygons to be processed as part of a single, spatially-consistent pipeline. However, such an approach still requires dedicated feature engineering and also ignores the spatial distribution and geometry details of individual geographic objects.
Another promising direction is the recently proposed location-aware self-supervised learning frameworks, including CSP \cite{Mai.2023b}, SatCLIP \cite{Klemmer.2025}, GeoCLIP \cite{Cepeda.2023}, Taxabind \cite{sastry2025taxabind}, RANGE \cite{Dhakal.2025}, and GAIR \cite{liu2026gair}, which perform cross-modal contrastive learning among neural representations of geospatially colocated geospatial observations (e.g., satellite images, ground-level images, geographic coordinates, geo-tagged texts, geo-tagged audios, etc). However, such kind of frameworks cannot deal with more complex geometry types such as polylines and polygons.

Thus, a unified SRL framework is needed, which can seamlessly handle diverse geospatially aligned data modalities within the same neural architecture while being able to preserve the fine-grained details (e.g., complex geometry details). 
Vector primitives can function as structured geospatial tokens analogous to linguistic tokens in vision-language architectures. Image tiles and geo-entities constitute dual input sequences to unified transformer backbones, with bidirectional cross-modal attention establishing natural spatial correspondences between appearance and semantics. Pretraining objectives orchestrate reconstruction of masked visual regions with prediction of occluded geo-entity attributes, exploiting complementary informational content across modalities. Generative geospatial frameworks further adapt pretrained language and vision models to spatiotemporal inference through knowledge-conditioned fine-tuning \cite{Wang.2024}, while domain-specific spectral foundation models provide robust visual encoders for unified architectures \cite{hong2024spectralgpt}.

This formulation naturally supports transfer across geospatial task families. Structural priors from vector tokens enhance perceptual boundary delineation and object consistency. Visual context disambiguates vector semantic attribution under environmental variability. Joint embeddings enable cross-modal geospatial retrieval and zero-shot reasoning over combined visual-structural evidence.

\subsection{Research Agenda for Unified Spatial Foundation Models}
We identify four key aspects for accelerating SRL research towards unified raster-vector spatial foundation models: (1) methodological innovations, (2) fairness in spatial foundation models, (3) evaluations that target both raster and vector-centric geospatial applications, and (4) approaches for uncertainty quantification and interpretability.

\paragraph{Methodological innovations:} Realizing unified spatial representation learning demands resolution of several architectural and empirical challenges. First, for the selection of model pretraining objectives, cross-modal alignment necessitates pretraining curricula that bidirectionally condition each modality upon the other. Joint contrastive objectives can be utilized to register pixel neighborhoods with corresponding geospatially aligned geo-entities, and reciprocal reconstruction losses can foster informational symbiosis. Beyond pretraining objectives, the integration mechanism between raster and vector representations presents another fundamental challenge. Due to their distinct spatial structures and semantic characteristics, simply passing raster and vector data through a shared encoder may not be the most effective solution. Instead, integration may occur at different stages of the learning pipeline depending on the downstream application. While joint encoding strategies represent one possible approach, raster and vector representations may also play complementary roles during decoding, reasoning, or generation stages. For example, raster observations may serve as contextual geo-priors for vector embeddings, whereas vector representations may provide semantic constraints or validation signals for raster-based predictions. Second, to capture the multi-scale and hierarchical nature of the geospatial data, multi-resolution spatial transformers must reconcile raster observations across octave-spaced scales with object-hierarchical vector representations. Hierarchical attention propagates contextual cues bidirectionally across representational granularity, balancing global geographic coherence with localized boundary fidelity. Third, in order to tackle spatial heterogeneity in a global setting, petabyte-scale dataset curation demands systematic spatial-temporal alignment between satellite archives and dynamic vector repositories, accounting for observational cadence disparities and regional completeness heterogeneity. Last but not least, to preserve the detailed information of each modality, such a unified SRL framework should support modality-specific encoders and decoders \cite{Mai.2024b} to support lossless data reconstruction. 

\paragraph{Fairness:} This curation must also account for \textit{Spatial Fairness} \cite{xie2022fairness} and the inherent imbalance in vector data density while mitigating geographic bias \cite{Wu.2024}. Unified models must learn to generalize from data-rich urban centers to data-sparse rural regions without introducing geographic bias \cite{Zhang.2026}. This is exacerbated by the fact that much raster data (e.g. optical or radar imagery) is available globally and largely unbiased, whereas vector data such as demographic information or crowdsourced OpenStreetMap data often exhibits large spatial coverage biases and largely favors high-income countries.

\paragraph{Evaluations:} Unified evaluation frameworks are essential for understanding and improving the capabilities of next-generation geospatial foundation models. While current standardized protocols such as PANGAEA \cite{Marsocci.2024}, REOBench \cite{XiangLi.2025}, and ExEBench \cite{ShanZhao.2025} have established vital baselines for raster-centric foundation models, and SRL frameworks like TorchSpatial \cite{Wu.2024} and RANGE \cite{Dhakal.2025} provide important primitives for location encoding, the field currently lacks a holistic benchmark for joint raster–vector reasoning. Future efforts must extend these existing evaluation paradigms to include cross-modal retrieval, vector-conditioned image segmentation, and visually-augmented spatial query resolution, while explicitly accounting for out-of-distribution (OOD) scenarios in open-world Earth observation \cite{GonzalezCalabuig.2025}. 

\paragraph{Uncertainty quantification \& interpretability:} Responsible geospatial artificial intelligence requires rigorous uncertainty propagation across modality fusion, as standard deep learning architectures often provide overconfident predictions on unrecognizable or OOD inputs \cite{GonzalezCalabuig.2025}. Building on foundational Bayesian approximations and deep ensemble methods \cite{Gawlikowski.2023}, new research must address the specific challenges of spatial uncertainty, including the identification of unknown environmental states and misclassified geospatial features. Furthermore, cross-modal interpretability and explanability mechanisms—such as those exploring representation uncertainty in Earth Observation \cite{GonzalezCalabuig.2025}—are essential to expose decision provenance and ensure that joint visual-structural reasoning remains trustworthy. Recent work suggests that sparse autoencoders can recover more monosemantic and interpretable geospatial concepts from learned Earth embeddings compared to raw representations; however, these findings remain preliminary and require further validation before they can support reliable scientific discovery or operational geospatial mapping applications. Ultimately, this research trajectory points toward the formalization of "Earth Embeddings" \cite{Klemmer.2025}—a multi-dimensional, AI-native representation of location that transcends the raster-vector dichotomy to provide a unified foundation for geospatial intelligence.

\section{Potentials of spatial representation learning beyond pixels} 
Unified SRL across raster and vector modalities could enable a new generation of geospatial AI applications by combining physical observations with structured semantic information. Such capabilities are particularly relevant for spatial reasoning tasks that require joint understanding of environmental and human systems.

First, unified SRL could enhance spatio-temporal monitoring of environmental change, socioeconomic indicators, and urban dynamics ~\citep{Liu.2025b}. For example, \citet{zheng2026satellite} introduced a satellite foundation model \textit{Tempov} pretrained on large-scale bi-temporal Landsat imagery pairs and adapted with sparse geolocated household survey labels, achieving better performance on dynamic wealth estimation than CNNs and general GeoFMs (e.g., \textit{Prithvi}~\citep{Szwarcman.2025}). In time-series remote sensing, satellite observations are frequently incomplete due to cloud occlusion, sensor limitations, or irregular revisit times ~\citep{fallah2026agflow}. Vector datasets containing semantic and structural attributes can provide conditioning signals ~\citep{fallah2025rareflow} for diffusion models and other generative frameworks to reconstruct missing observations or enforce temporal consistency. These approaches may improve the quality and physical plausibility of generated outputs, particularly in rapidly changing urban environments.

Second, unified SRL can improve geolocalization of heterogeneous data sources such as satellite imagery, street-view imagery, camera traps, and social media posts ~\citep{li2022geoai}. In disaster mapping and damage assessment, building footprints and semantic attributes can guide decoding and segmentation, improving discrimination of functionally different but visually similar structures that may exhibit different vulnerability or damage patterns. Additional modalities such as ground sensor observations, POIs, and geo-referenced texts, could further validate and improve the interpretability of geospatial representations~\citep{Liu.2025b}. One challenge is how to make accurate predictions from the unified neural embeddings with the coexistence of spatial, temporal, vector, and spectral resolution gaps as well as mitigating and flagging hallucinations from ingested generative sources. 

Third, the harmonization of geospatial vector and raster datasets in unified SRL would enhance the spatial capabilities of ``world models''~\citep{hao2023reasoning}, which enable AI agents to reason about 3D environments and spatial relationships (distance, direction, and topology) and act under physical constraints. By combining pixel-based visual observations or generations with object-based semantics, these agents can perform macro-level spatial reasoning and planning-by-prediction with world models and apply in autonomous driving and robotics~\citep{dong2026learning}. Some of the major challenges include the misalignment between representation learning metrics and downstream task objectives and the lack of explicit modeling of causal structures, which may exaggerate errors and prediction drift.

\section{Conclusion} 
Raster and vector representations furnish complementary vantages upon geographic reality: continuous perceptual density confronts discrete structural semantics. Contemporary geospatial artificial intelligence processes these modalities through disjointed, information-degrading transformations that undermine representational coherence. Native unified spatial learning synthesizes both domains within shared architectures, harnessing cross-modal attention and collaborative pretraining to forge contextually grounded geospatial intelligence.

This perspective delineates the conceptual foundations from raster perception through vector structure, scrutinizes limitations of prevailing integration paradigms, and charts a precise research trajectory toward unified spatial foundation models. Through perceptual-structural synthesis, subsequent geospatial systems promise enhanced robustness, interpretability, and real-world utility across environmental monitoring, urban systems analysis, emergency response, and beyond. The research community confronts an unambiguous imperative: to construct spatial artificial intelligence that comprehends both Earth's observable surface and the deliberate structure humanity imposes upon it.


{
    \small
    \bibliographystyle{ieeenat_fullname}
    \bibliography{Spatial_Representation_Learning}
}

\end{document}

%% file: preamble.tex
\usepackage{verbatim}

\usepackage{lipsum}

\usepackage{tcolorbox}
\usepackage{xcolor}

\usepackage{tcolorbox}
\usepackage{xcolor}